%% file: pakdd.tex
\documentclass[runningheads,a4paper]{llncs}

\usepackage{graphicx}
\usepackage{times}
\usepackage{helvet}
\usepackage{courier}
\usepackage{epsfig}
\usepackage{amssymb}
\usepackage{amsmath}
\usepackage{amsfonts}
\usepackage{algorithm}
\usepackage{algorithmic}
\usepackage{mdwlist}
\usepackage{multirow}
\usepackage{color}
\usepackage{subfigure}
\usepackage{url}
\usepackage{enumerate}
\usepackage{threeparttable}

\usepackage{url}
\urldef{\mailsa}\path|{alfred.hofmann, ursula.barth, ingrid.haas, frank.holzwarth,|
\urldef{\mailsb}\path|anna.kramer, leonie.kunz, christine.reiss, nicole.sator,|
\urldef{\mailsc}\path|erika.siebert-cole, peter.strasser, lncs}@springer.com|    
\newcommand{\keywords}[1]{\par\addvspace\baselineskip
\noindent\keywordname\enspace\ignorespaces#1}

\begin{document}

\mainmatter  

\title{Marked Temporal Dynamics Modeling based on Recurrent Neural Network}

\titlerunning{Marked Temporal Dynamics Modeling based on Recurrent Neural Network}

%
%
\author{Yongqing Wang, Shenghua Liu, Huawei Shen, Xueqi Cheng}
\authorrunning{}

\institute{CAS Key Laboratory of Network Data Science and Technology, \\ Institute of
Computing Technology, Chinese Academy of Sciences, Beijing 100190, China}

%
%

\toctitle{Marked Temporal Dynamics Modeling based on Recurrent Neural Network}
\tocauthor{}
\maketitle

\begin{abstract}
\input{0-abs.tex}
\end{abstract}

\input{1-intro.tex}

\input{2-model.tex}
\input{3-opt.tex}
\input{4-exp.tex}
\input{5-conclusion.tex}

\bibliographystyle{splncs03}
\bibliography{social}

\end{document}

%% file: 0-abs.tex

We are now witnessing the increasing availability of event stream data, i.e., a
sequence of events with each event typically being denoted by the time it
occurs and its mark information (e.g., event type). A fundamental problem is to
model and predict such kind of marked temporal dynamics, i.e., when the next
event will take place and what its mark will be. Existing methods either
predict only the mark or the time of the next event, or predict both of them,
yet separately. Indeed, in marked temporal dynamics, the time and the mark of
the next event are highly dependent on each other, requiring a method that
could simultaneously predict both of them. To tackle
this problem, in this paper, we propose to model marked temporal dynamics by
using a \emph{mark-specific} intensity function to explicitly capture the
dependency between the mark and the time of the next event. Extensive
experiments on two datasets demonstrate that the proposed method outperforms
state-of-the-art methods at predicting marked temporal dynamics.

\keywords{marked temporal dynamics, recurrent neural network, event stream
data}

%% file: 1-intro.tex
\section{Introduction}
\label{sec:intro}

There is an increasing amount of event stream data, i.e. a sequence of events
with each event being denoted by the time it occurs and its mark information
(e.g. event type). Marked temporal dynamics offers us a way to describe this
data and potentially predict events. For example, in microblogging platforms,
marked temporal dynamics could be used to characterize a user's sequence of
tweets containing the posting time and the topic as mark~\cite{gao2015modeling};
in location based social networks, the trajectory of a user gives rise to a
marked temporal dynamics, reflecting the time and the location of each
check-in~\cite{liuAAAI2016}; in stock market, marked temporal dynamics
corresponds to a sequence of investors' trading behaviors, i.e., bidding or
asking orders, with the type of trading as mark~\cite{cao2010detecting}; 
An ability to predict marked temporal
dynamics, i.e., predicting when the next event will take place and what its
mark will be, is not only fundamental to understanding the regularity or
patterns of these underlying complex systems, but also has important
implications in a wide range of applications, from viral marketing and traffic
control to risk management and policy making.

Existing methods for this problem fall into three main paradigms, each with
different assumptions and limitations. The first category of methods focuses on
predicting the mark of the next event, formulating the problem as a
discrete-time or continuous-time sequence prediction
task~\cite{isaacson1976markov,tankov2003financial}.
These methods gained success at modeling the transition probability across
marks of events. However, they lack the power at predicting when the next event
will occur. 

The second category of methods, on contrary, aims to predict when
the next event will occur~\cite{gomez2013modeling}. These
methods either exploit temporal correlations for
prediction~\cite{pinto2013using,Szabo2010Commun} or conduct
prediction by modeling the temporal dynamics using certain temporal process,
such as self-exciting Hawkes
process~\cite{CranePNAS08,bao2015modeling},
various Poisson process~\cite{shen2014modeling,gao2015modeling} , and other
auto-regressive
processes~\cite{engle1998autoregressive,vaz2013self}.
These methods have been successful used in modeling and predicting temporal
dynamics.
However, these models are unable to predict the mark. 

Besides the above two categories of
methods, researchers recently attempt to directly model the marked temporal
dynamics~\cite{gunawardana2016universal}. A recent work~\cite{DuKDD2016} 
used recurrent neural network to automatically learn history embedding, and then
predict both, yet separately, the time and the mark of the next event. This work
assumes that time and mark are independent on each other given the historical
information. Yet, such assumption fails to capture the
dependency between the time and the mark of the next event. For example,
when you have lunch is affected by your choice on restaurants, since
different restaurants imply difference in geographic distance
and quality of service.
The separated
prediction by maximizing the probability on mark and time does not imply the most likely event.
In sum, we still lack a model that
could consider the interdependency of mark and time when predicting the next
event.

In this paper, we propose a novel model based on recurrent neural network (RNN),
named RNN-TD, to capture the dependence between the mark of an event and its
occurring time. The key idea is to use a mark-specific intensity function to
model the occurring time for events with different marks. 
Besides, RNN can help to relieves the disscussion of the explicit dependency
structure among historical events, which embeds sequential characteristics.
The benefits of our proposed model are
three-fold: 1) It models the mark and the time of the next event
simultaneously; 2) The mark-specific intensity function explicitly captures the
dependency between the occurring time and the mark of an event; 3) The
involvement of RNN simplifies the modeling of depenedency on historical events.


We evaluate the proposed model by extensive experiments on large-scale real
world datasets from Memetracker\footnote{http://www.memetracker.org} and
Dianping\footnote{http://www.dianping.com}.
Compared with several state-of-the-art methods, RNN-TD outperforms
them at prediction of marks and times. We also conduct case study
to explore the capability of event prediction in RNN-TD. The experimental
results prove that it can better model marked temporal dynamics.

%% file: 2-model.tex
\section{Proposed Model}
\label{sec:model}


In this paper, we focus on the problem of modeling marked temporal dynamics.
Before diving into the details of the proposed model, we first clarify two
main motivations underlying our model. 

\subsection{Motivation}
\begin{figure}[t]
\centering
\subfigure[] {
\label{fig:motivations}
\includegraphics[width=0.35\textwidth]{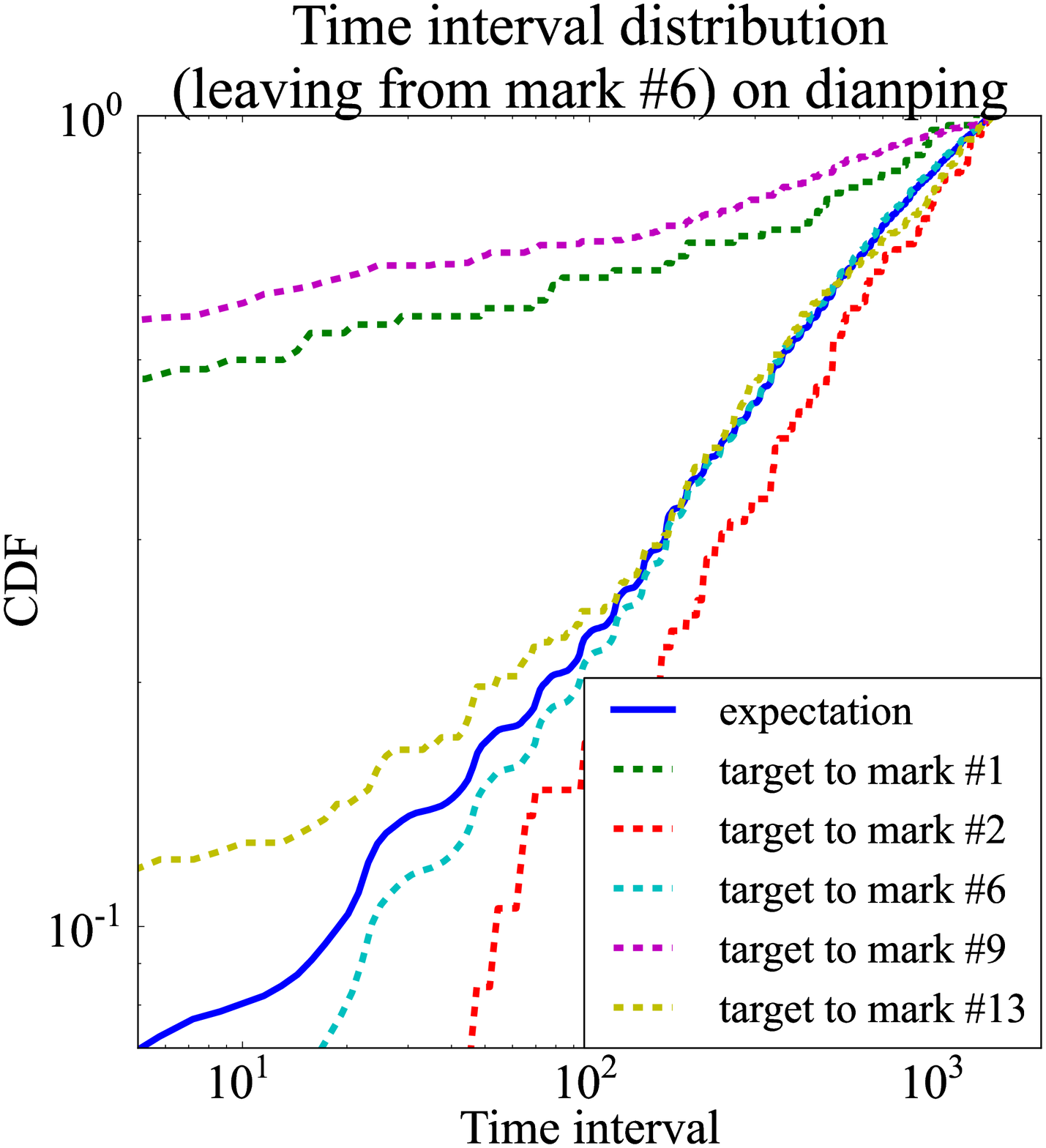}
}
\subfigure[] {
\includegraphics[width=0.55\textwidth]{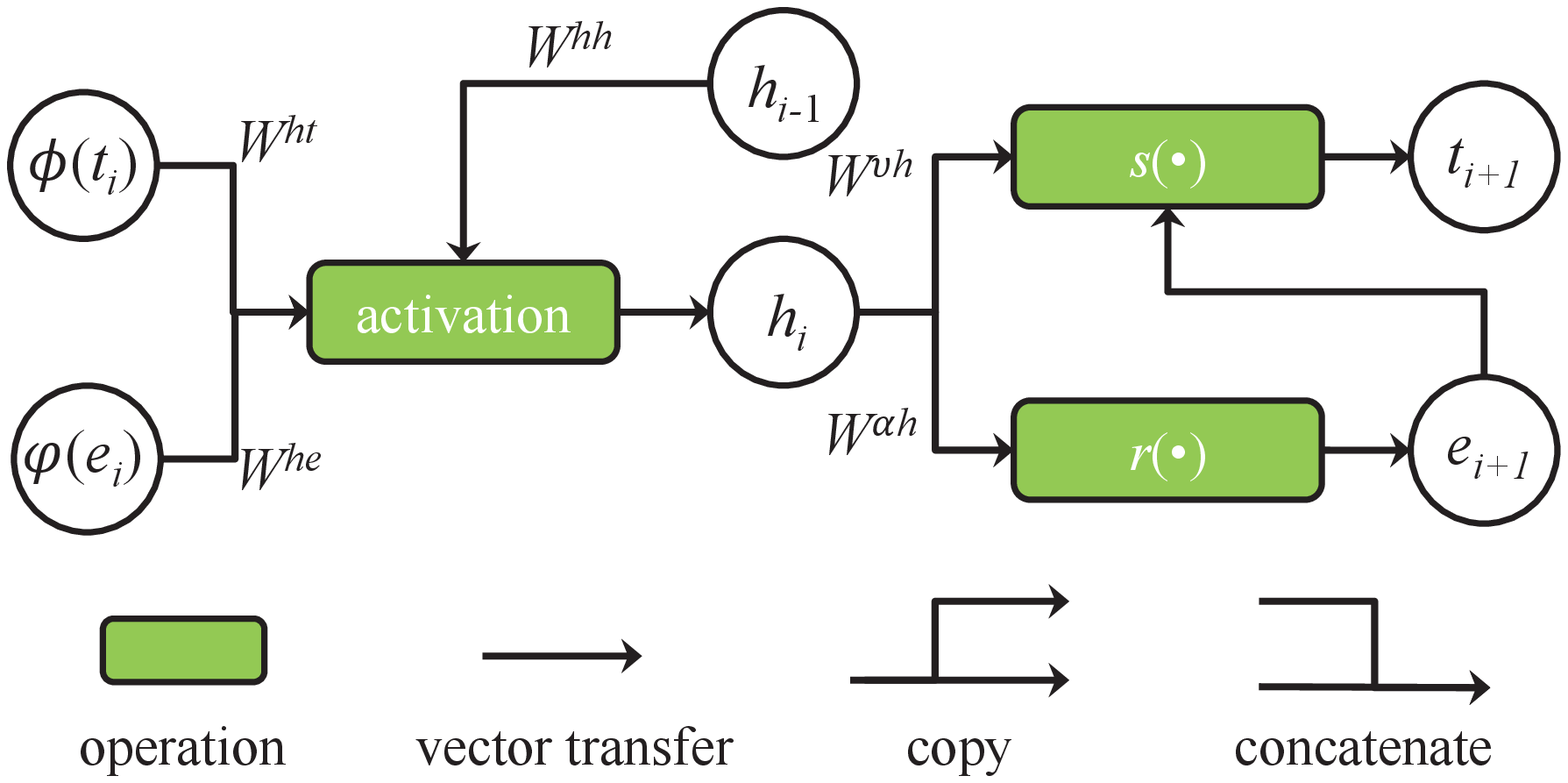}
\label{fig:rnn-td}
}
\caption{(a) High
variance existed in time interval distribution when targeting to different
marks. (b) The architecture of RNN-TD. Given the event sequence
$S=\{(t_i,e_i)\}_{i=1}$, the $i$-th event $(t_i,e_i)$
is mapped through function $\phi(t)$ and $\varphi(e)$ into vector spaces as
inputs in RNN. Then the inputs $\phi(t_i)$ and $\varphi(e_i)$ associated with
the last embedding $h_{i-1}$ are fed into hidden units in order to update $h_i$.
Dependent on embedding $h_i$, RNN-TD outputs the next event type $e_{i+1}$ and
correspondint time $t_{i+1}$.}
\end{figure}

In real scenarios, mark and time of next event are highly dependent on each
other. 
To validate this point, we focus on a practical case in Dianping. We extract the
trajectories starting from the same location (mark\#6) and get the statistic
results to examine if the time interval between two consecutive events are
discriminative to each other with respect to different marks. The statistics of
time interval distribution are represented in Fig.~\ref{fig:motivations}.
We can observe that large variance exists in the distributions
when consumers make different choices. It motivates us to model mark-specific
temporal dynamics.




Second, existing
works~\cite{isaacson1976markov,tankov2003financial}
attempted to formulate marked temporal dynamics by Markov random processes
with varying orders. However, the generation
of next event requires strong prior knowledge on dependency of history. Besides, long dependency on history causes
state-space explosion problem in practice. Therefore, we propose a RNN-based
model which learns the dependency by deep structure. It embeds history
information into vectorized representation when modeling
sequences.
The generation of next event is only dependent on history embedding.


\subsection{Problem Formulation}
An event sequence
$S=\{(t_i, e_i)\}_{i=1}$ is a set of events in ascending order of
time. The tuple $(t_i, e_i)$ records the $i$-th event
in the sequence $S$, and the variables $t_i\in
\mathcal{T}$ and $e_i\in \mathcal{E}$ denote the time and the mark
respectively, where $\mathcal{E}$ is a countable state space including all
possible marks and $\mathcal{T}\in \mathbb{R}^+$ is the time space in which
observed marks take place. We could have various instantiation in different
applications. 


Then the likelihood of observed sequence $S$ can be unfolded by chain rule as
follows,
\begin{equation*}
\label{eq:jump_pcs_lkhd}
\small
P\left(S\right)=\prod_{i=1}^{|S|}p(t_i,e_i|H_{t_i}),
\end{equation*}
where $H_{t_i}=\{(t_l,e_l)| t_l< t_i, e_l\in\mathcal{E}\}$ refers to all
related historical events occurring before $t_i$. In practice, the joint
probability of a pair of mark and time can be written by Bayesian rule as
follows
\begin{equation} 
\label{eq:joint_process}
p(t_i,e_i|H_{t_i})=r(e_i|H_{t_i})s(t_i|e_i,H_{t_i}),
\end{equation}
where $r(e_i|H_{t_i})$ is the transition probability related to $e_i$ and
$s(t_i|e_i,H_{t_i})$ is the probability distribution function of time given a
specific mark. 


Then we propose a general model to parameterize $r(e_i|H_{t_i})$ and
$s(t_i|e_i,H_{t_i})$ in marked temporal dynamics modeling, named RNN-TD.
Recurrent neural network (RNN) is a feed-forward neural network for modeling sequential data.
In RNN, the current inputs are fed into hidden units by nonlinear
transformation, jointly with the outputs from the previous hidden units.
The feed-forward architecture is replicative in both inputs and outputs
so that the representation of hidden units is dependent on not only
current inputs but also encoded historicial information. The adaptive size of
hidden units and nonlinear activation function (e.g., sigmoid, tangent
hyperbolic or rectifier function) make neural network capable of approximating
arbitrary complex function in huge function space~\cite{barzel2015constructing}.

The architecture
of RNN-TD is depicted in Fig.~\ref{fig:rnn-td}. The inputs of a event
$(t_i,e_i)$ is vectorized by mapping function $\phi(\cdot)$ and
$\varphi(\cdot)$.
Then the $i$-th inputs associated with the last embedding $h_{i-1}$ are fed into hidden
units in order to update $h_i$. Given the $i$-th event $(t_i, e_i)$, the embedding
$h_{i-1}$ and mapping functin $\phi$ and $\varphi$, the representation of hidden units in
RNN-TD can be calculated as
\begin{equation}
h_i=\sigma\left(W^{ht}\phi(t_i)+W^{he}\varphi(e_i)+W^{hh}h_{i-1}\right),
\end{equation}
where $\sigma$ is the activation function, and $W^{ht}$, $W^{he}$ and $W^{hh}$
are weight matrices in neural network.
The procedure is iteratively executed until the end of sequence. Thus, the
embedding $h_i$ encodes the $i$-th inputs and the historical context $h_{i-1}$. 

Based on the history embedding $h_i$, we can derive the probability of the
$(i+1)$-th event in an approximative way,
\begin{equation} 
p(t_{i+1},e_{i+1}|H_{t_{i+1}})\approx p(t_{i+1},e_{i+1}|h_{i})=r(e_{i+1}|h_{i})s(t_{i+1}|e_{i+1},h_{i}).\\
\end{equation}

Firstly we formalize the conditional transition probability
$r(e_{i+1}|h_{i})$.
The conditional transition probability can be derived by a softmax
function which is commonly used in neural network for parameterizing categorical
distribution, that is,
\begin{equation}
\label{eq:trans_prob}
r(e_{i+1}|h_{i})=\frac{\exp\left(W_{k}^{\alpha h} h_i\right)}{\sum_{j=1}^K
\exp\left(W_{j}^{\alpha h} h_i\right)},
\end{equation}
where row vector $W_{k}^{\alpha h}$ is $k$-th row of weight matrix indexed by
the mark $e_{i+1}$.

Then we consider the probability distribution function
$s(t_{i+1}|e_{i+1},h_{i})$. The probability distribution function describes
\textit{the observation that nothing but mark $e_{i+1}$ occurred until time
$t_{i+1}$ since the last event}. We define a random variable $T_e$ about
occuring time of next event with respect to mark $e$, and the probability distribution function
$s(t_{i+1}|e_{i+1},h_{i})$ can be formalized as
\begin{equation}
\label{eq:mark_lkhd}
s(t_{i+1}|e_{i+1},h_{i})=P(T_{e_{i+1}}=t_{i+1}|e_{i+1},h_{i})
\prod_{e\in\mathcal{E}\setminus e_{i+1}}P(T_e> t_{i+1}|e_{i+1}, h_{i}),
\end{equation}
where the probability $P(T_e>t_{i+1}|e_{i+1}, h_{i})$ depicts that the
occuring time of event with mark $e$ is out of the range $[0, t_{i+1}]$, and
$P(T_{e_{i+1}}=t_{i+1}|e_{i+1},h_{i})$ is the conditional probability density
function representing the fact that mark $e_{i+1}$ is ocurring until time $t_{i+1}$. 

To formalize the Eq.~(\ref{eq:mark_lkhd}), we define
\textit{mark-specific conditional intensity function} as
\begin{equation}
\label{eq:mark_lambda}
\lambda_{e}(t_{i+1})=\frac{f_e(t_{i+1}|e_{i+1}, h_{i})}{1-F_e(t_{i+1}|e_{i+1},
h_{i})},
\end{equation}
where $F_e(t_{i+1}|e_{i+1}, h_{i})$ is the cumulative distribution function of
$f_e(t_{i+1}|e_{i+1}, h_{i})$. 
According to Eq.~(\ref{eq:mark_lambda}), we can derive the cumulative
distribution function
\begin{equation}
\label{eq:cum_dist}
F_e(t_{i+1}|e_{i+1}, h_{i})=1-\exp(-\int_{t_{i}}^{t_{i+1}}\lambda_{e}(\tau)
d\tau).
\end{equation}
The probability of $P(T_e>t_{i+1}|e_{i+1}, h_{i})=1-F_e(t_{i+1}|e_{i+1},
h_{i})$.
Then we can derive the mark-specific conditional probability
density function by Eq.~(\ref{eq:cum_dist}) as
\begin{equation}
\label{eq:ms_dens_func}
\small
P(T_{e}=t_{i+1}|e_{i+1},h_{i})=f_e(t_{i+1}|e_{i+1},
h_{i})=\lambda_{e}(t_{i+1})\exp(-\int_{t_{i}}^{t_{i+1}} \lambda_{e}(t)dt).
\end{equation}
Substituting Eq.~(\ref{eq:cum_dist}) and Eq.~(\ref{eq:ms_dens_func}) into
the likelihood of Eq.~(\ref{eq:mark_lkhd}), we can get
\begin{equation}
\label{eq:mark_survival}
s(t_{i+1}|e_{i+1},h_{i})=\lambda_{e_{i+1}}(t_{i+1})\exp(-\int_{t_{i}}^{t_{i+1}}
\lambda(t)dt),
\end{equation}
where $\lambda(\tau)=\sum_{e\in\mathcal{E}}\lambda_e(\tau)$ is the summation of
all conditional intensity function. 

The key to specify probability distribution function $s(t_{i+1}|e_{i+1},h_i)$ is
parameterization of mark-specific conditional intensity function $\lambda_{e}$.
We parameterize $\lambda_{e}$ conditioned on $h_i$ as follows,
\begin{equation}
\label{eq:rnntd_lambda}
\lambda_{e}(t)=\nu_{e}\cdot
\tau(t;t_{i})=\exp\left(W_{k}^{\nu h} h_i\right)\tau(t;t_{i}),
\end{equation}
where row vector $W_{k}^{\nu h}$ denotes to the $k$-th row of weight matrix
corresponding to mark $e$. In Eq.~(\ref{eq:rnntd_lambda}), the mark-specific
conditional intensity function is splited into two parts:
$\nu_{e}=\exp(W_{j'}^{\nu h} h_i)$ is a nonnegative scalar as the
constant part with respect to time $t$, and $\tau(t;t_{i})\geq 0$ refers to
an arbitrary time shaping function~\cite{gomez2013modeling}. For simplicity, we
consider two well-known parametric models for time shaping function: exponential
and constant, i.e., $\exp(wt)$ and $c$.

At last, given a collection of event sequences $\mathcal{C}=\{S_m\}_{m=1}^N$, we
suppose that each event sequence $S_m$ is independent of others. 
As a result, the logarithmic likelihood of a set of event sequences is the
sum of the logarithmic likelihood of the individual sequence. Given the
source of event sequence, the negative logarithmic likelihood of the set of
event sequences $\mathcal{C}$ can be estimated as,
\begin{equation*}
\small
\begin{aligned}
\mathcal{L}\left(\mathcal{C}\right)=-\sum_{m=1}^N\sum_{i=1}^{|S_m|-1}\Bigg[
&W_{k}^{\alpha h} h_i-\log\sum_{j=1}^K \exp\left(W_{j}^{\alpha h} h_i\right)\\
&+W_{k}^{\nu h} h_i+\log\tau(t;t_{i})-\sum_{e\in\mathcal{E}}\exp\left(W_{j'}^{\nu h}
h_i\right)\int_{t_{i}}^{t_{i+1}}\tau(t;t_i)dt\Bigg].\\
\end{aligned}
\end{equation*}
In addition, we want to induce sparse structure in vector $\nu$ in order
that not all event types are available to be activated based on $h_i$.
For this
purpose, we introduce lasso regularization on $\nu$, i.e.,
$\|\nu\|_1$~\cite{tibshirani1996regression}.
Overall, we can learn parameters of RNN-TD by minimizing the
negative logarthmic likelihood
\begin{equation}
\arg\min_{W} \mathcal{L}(\mathcal{C})+\gamma\|\nu\|_1,
\end{equation}
where $\gamma$ is the trade-off parameter. 


As last, we can estimate the next most likely events in two steps by RNN-TD:
1) estimate the time of each mark by expectation
$t_{i+1}=\int_{t_i}^{\infty}t\cdot s(t|e_{i+1},h_{i})dt$; 2) calculate the
likelihood of events according to the mark-specific expectation time, and then
rank events in descending order of likelihood.

%% file: 3-opt.tex
\section{Optimization}
\label{sec:opt}

In this section, we introduce the learning process of RNN-TD. We apply
back-propagation through time (BPTT)~\cite{chauvin1995backpropagation} for
parameter estimation. With BPTT method, we need to unfold the neural network in consideration of
sequence size $|S_m|$ and update the parameters once after the completed forward
process in sequence. 
We employ Adam~\cite{kingma2015method}, an efficient
stochastic optimization algorithm, with mini-batch techniques to iteratively update all
parameters. We also apply early stopping method~\cite{prechelt1998automatic}
to prevent overfitting in RNN-TD. The stopping criterion is achieved when
the performance has no more improvement in validation set. The mapping function of $\phi(t)$
is defined by temporal features associated with $t$, e.g., logarithm time
interval $\log(t_i-t_{i-1})$ and discretization of numerical attributes on
year, month, day, week, hour, mininute, and second. Besides, we employ
orthogonal initialization method~\cite{henaff2016orthogonal} for RNN-TD in order
to speed up convergence in training process. The embedding learned by
word2vec~\cite{MikolovNIPS2013,perozzi2014deepwalk} is used to initialize the
parameter of mapping function $\varphi(e)$. The good initialization provided 
by the embedding can speed up convergence for
RNN~\cite{mikolov2013efficient}.

%% file: 4-exp.tex
\section{Experiments}
\label{sec:exp}
Firstly, we
introduce baselines, evaluation metrics and datasets of our experiments.
Then we conduct experiments on real data to validate the performance of RNN-TD in
comparison with baselines. 

\subsection{Baselines}
Both mark prediction and time prediction are evaluated, and the following models
are chosen for comparisons in the two prediction tasks.

(1) Mark sequence modeling.
\begin{itemize}
  \item \textbf{MC:} The markov chain model is a classic sequence modeling
  method. We compare with markov chain of varying orders from one to three,
  denoted as MC1, MC2 and MC3.
  \item \textbf{RNN:} RNN is a state-of-the-art method for discrete time
  sequence modeling, successfully applied in language model. To fairly
  justify the performance between RNN and our proposed method,
  We use the same inputs in both RNN and RNN-TD.
\end{itemize}

(2) Temporal dynamics modeling. 
We choose point processes and mark-specific point processes with different
characterizations as baselines.
\begin{itemize} 
  \item \textbf{PP-poisson:} The intensity function related to mark is
  parameterized by a constant, depicting the leaving rate from last event.
  \item \textbf{PP-hawkes:} The intensity function related to mark $e$ is
  parameterzied by
  \begin{equation} 
  \label{eq:hawke_intens_func}
  \lambda(t;e)=\lambda(0;e)+\alpha\sum_{t_i<t}\exp\left(-\frac{t-t_i}{\sigma}\right),
  \end{equation}
  where $\sigma=1$ and $\lambda(0;e)$ is a intrinsic rate defined on mark $e$
  when $t=0$.
  \item \textbf{MSPP-poisson:} We define the mark-specific intensity function
  by a parametric matrix, depicting the rate from one mark to another. 
  \item \textbf{MSPP-hawkes:} The mark-specific intensity function is
  parameterized by Eq.~(\ref{eq:hawke_intens_func}) where the constant rate is
  specialized according to mark pairs in parametric matrix.
\end{itemize}

We also compare with the model that has the ability to generate both mark and
temporal sequences. 
\begin{itemize} 
  \item \textbf{RMTPP:} Recurrent marked temporal point process
  (RMTPP)~\cite{DuKDD2016} is a method which independently models both mark and
  time information based on RNN.
\end{itemize}


\subsection{Evaluation Metrics}
Serveral evaluation metrics are used when measuring the performance in mark
prediction and time prediction tasks. We regard the mark
prediction task as a ranking problem with respect to
transition probability. The prediction performance is evaluated by \textit{Accuracy} on top
$k$ (Acc@$k$) and \textit{Mean Reciprocal Rank} (MRR)~\cite{voorhees1999trec}.
On time prediction task, we define tolerance $\theta$ over the prediction error
between estimated time and practical occuring time. The prediction accuracy on
time prediction with respect to tolerance $\theta$ is formulated as,
\begin{equation*}
\small
\text{Acc@}\theta=\frac{\sum_{m=1}^N\sum_{i=1}^{|S_m|-1}\delta\left(|E(t;e_{i+1},h_i)-t_{i+1}|<\theta\right)}{\sum_{m=1}^N
(|S_m|-1)},
\end{equation*}
where $\delta$ is an indicator function. Larger scores in Acc@$k$, MRR and
Acc@$\theta$ indicate better predictions. 

\subsection{Datasets}
We conduct experiments on two real datasets from two different
scenarios to evaluate the performance of different methods:

\begin{table}[t]
\centering
\small
\caption{Performance of mark prediction on two datasets}
\label{tb:mrr_real_data}
\begin{tabular*}{.85\textwidth}{cccccccc}
\hline
&  & MRR & Acc@1 & Acc@3 & Acc@5 & Acc@10 & Acc@20 \\
\hline
\multirow{9}{*}{Memetracker} & MC1 & 0.4634 & 0.2948 & 0.4595 & 0.6659 &
0.8253 & 0.9209
\\
& MC2 & 0.4788 & 0.3155 & 0.4706 & 0.6773 & 0.8301 & 0.9186\\
& MC3 & 0.4670 & 0.3149 & 0.4583 & 0.6550 & 0.7891 & 0.8619\\
& RNN & 0.4780 & 0.3202 & 0.4746 & 0.6825 & 0.8315 & 0.9201\\
& RMTPP & 0.4833 & 0.3241 & 0.4834 & 0.6926 & 0.8386 & 0.9267\\
& RNN-TD(c) & 0.4820 & 0.3220 & 0.4790 & 0.6895 & 0.8393 & 0.9270\\
& RNN-TD(exp) & 0.4849 & 0.3266 & 0.4835 &
0.6929 & 0.8400 & 0.9273
\\
& RNN-TD*(c) & 0.4820 & 0.3220 & 0.4790 & 0.6895 & 0.8393 & 0.9270\\
& RNN-TD*(exp) & \textbf{0.4851} & \textbf{0.3266} & \textbf{0.4844} &
\textbf{0.6937} & \textbf{0.8407} & \textbf{0.9274}
\\
\hline \multirow{9}{*}{Dianping} & MC1 &
0.6174 & 0.5231 & 0.6157 & 0.7212 & 0.7963 & 0.8787 
\\
& MC2 & 0.6260 & 0.5280 & 0.6396 & 0.7393 & 0.8007 & 0.8513 \\
& MC3 & 0.5208 & 0.4462 & 0.5395 & 0.6035 & 0.6332 & 0.6569  \\
& RNN & 0.6355 & 0.5123 & 0.6135 & 0.7153 & 0.7905 & 0.8656 \\
& RMTPP & 0.6620 & 0.5482 & 0.6554 & 0.7578 & 0.8271 & 0.8935 \\
& RNN-TD(c) & 0.6663 & 0.5524 & 0.6601 & 0.7628 &
0.8346 &
0.8999
\\
& RNN-TD(exp) & 0.6635 & 0.5448 & 0.6560 & 0.7638 &
0.8345 & 0.8988
\\
& RNN-TD*(c) & \textbf{0.6663} & \textbf{0.5524} & \textbf{0.6602} & 0.7628 &
0.8346 &
\textbf{0.8999}
\\
& RNN-TD*(exp) & 0.6635 & 0.5452 & 0.6566 & \textbf{0.7641} &
\textbf{0.8351} & 0.8990
\\
\hline
\end{tabular*}
\begin{minipage}{.85\textwidth}
p.s. the experimental results from * are dependent with given time. 
\end{minipage}
\end{table}

\begin{itemize}
  \item \textbf{Memetracker\cite{LeskovecKDD2009}:} Memetracker corpus contains
  articles from mainstream media and blogs from August 1 to October 31, 2008 with about 1
    million documents per day. Contents in the corpus are organized according to
  	topics by the proposed method in~\cite{LeskovecKDD2009}. We use top 165
  	frequent topics and organize the posting sequence about posted blogs and
  	post-time by users. The whole posting sequence of each user is splited into
  	parts as follows, 1) get the statistics of time intervals between two
  	consecutive posted blogs, 2) empirically estimate the period of user's
  	posting behavior, 3) and divide the whole sequence into several parts
  	according to the estimated period. To avoid processing short sequences, we
  	also ignore the sequences whose length are less than 3. The processed dataset
  	contains 1,481,491 posting sequences, and the time interval between two
  	consecutive blogs is ranged from $2.77\times 10^{-4}$ to 99.68 hours.
  \item \textbf{Dianping:} Dianping provide an online restaurant rating service
    in China, including coupon sales, bill payment, and reservation. We extract
	transaction coupon sales from top 256 popular stores located in Xidan
	bussiness district of Beijing from year 2011 to 2015. The consumption
	sequences of users are divided into segments as the same steps done in memetracker.
    Because of the existence of sparse shopping records in users, we also limit
    that time interval between two consecutive consumptions is two months. The
    processed dataset contains 221,893 event sequences, and the time interval
    between two consecutive consumptions is ranged from $2.77\times 10^{-4}$ to 1440 hours.
\end{itemize}

On both datasets, we randomly pick up 80\% of sequence data as training set, and
the rest data are divided into two parts equally as validation
set and test set respectively.

\begin{figure}[t]
\centering
\subfigure[Experiments on Memetracker] {
\label{fig:exp_memetracker}
\includegraphics[width=0.43\textwidth]{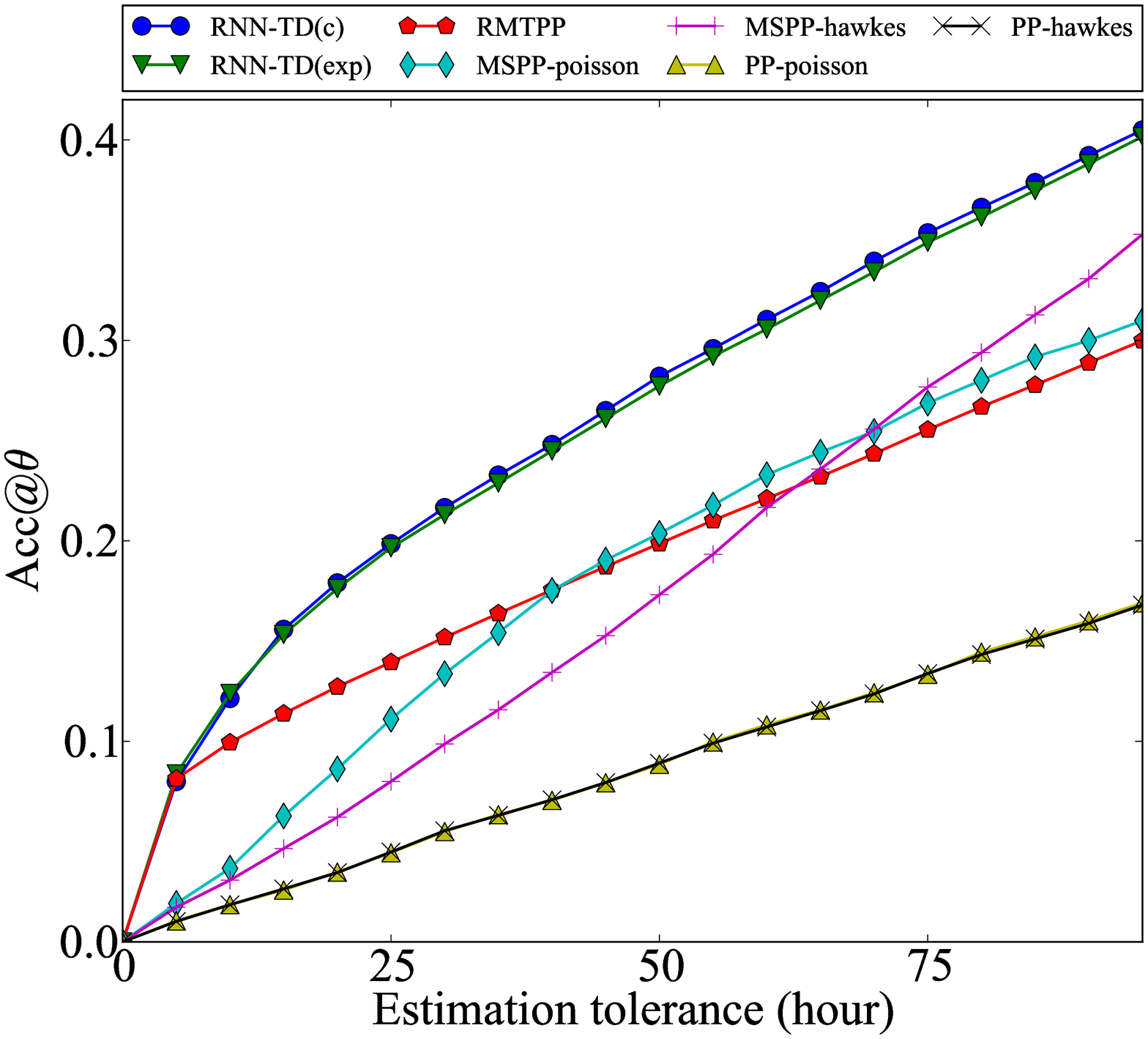}
}
\subfigure[Experiments on Dianping] {
\label{fig:exp_dianping}
\includegraphics[width=0.43\textwidth]{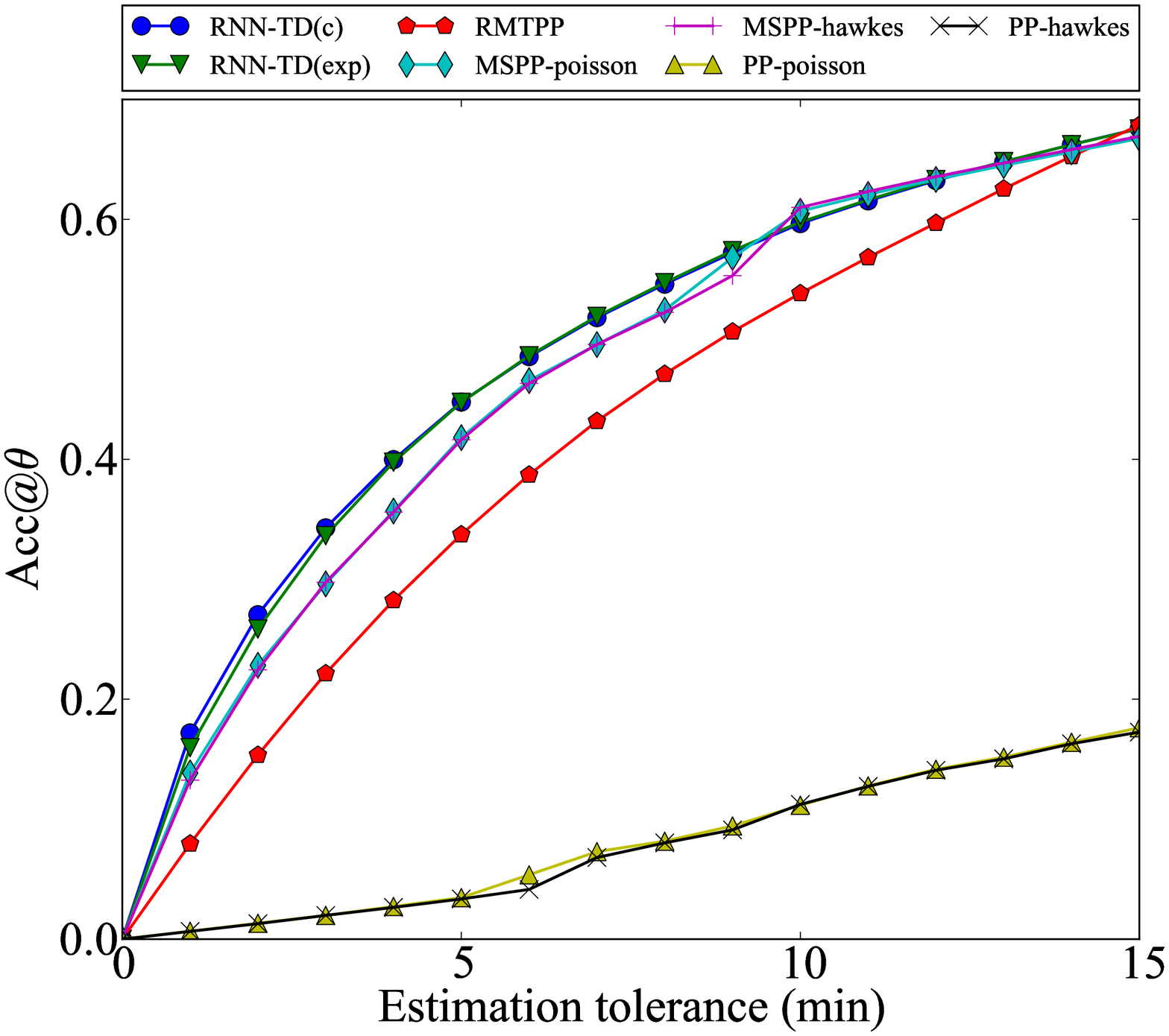}
}
\caption{Performance of timing prediction on two datasets.}
\label{fig:exp_real_data}
\end{figure}

\subsection{Performance of Mark Prediction}

The performance of mark prediction is evaluated using metrics Acc@$k$ and MRR.
The experimental results are shown in Table~\ref{tb:mrr_real_data}. Comparing
with MC1, MC2, MC3 and RNN, RNN-TD(c) and RNN-TD(exp) achieve significant
improvements over all metrics in both datasets. In Memetracker, RNN-TD(exp)
outperforms RMTPP in MRR at significance level of 0.1, and achieve a
little improvements than RMTPP in Acc@1,3,5,10 and 20. However, the performance of
RNN-TD(c) is worse than RMTPP. In Dianping, RNN-TD(c) achieves 
improvements than RMTPP in metrics of MRR and Acc@5 at significance level of 0.1
and metrics of Acc@10 and Acc@20 at significance level of 0.01. Besides,
RNN-TD(exp) achieves improvements than RMTPP in metrics of Acc@20
at significance level of 0.1 and metrics of Acc@5 and Acc@10 at significance
level of 0.01. The experimental results indicate that RNN-TD can better learn
the mark generation by jointly optimizing mark-specific conditional intensity
function with respect to different time shaping function applied in tasks.
 
We also conduct experiments according to event likelihood on RNN-TD with the
given time, marked as RNN-TD*.
The results of RNN-TD*(exp) performs little better than RNN-TD(exp) over all
metrics in both datasets, However, the performance of RNN-TD*(c) is almost the same as
RNN-TD(c). It demonstrates the robustness of RNN-TD on mark prediction whether
or not given the occuring time. Besides, RNN-TD with exponential form of time
shaping function has larger effects on given time than the constant form.

\subsection{Performance of Time Prediction}
We evaluate the performance of time prediction by Acc@$\theta$.
The predictions of RNN-TD and MSPP are based on true marks.
Fig.~\ref{fig:exp_memetracker} and Fig.~\ref{fig:exp_dianping} show the
experimental results of RNN-TD and baselines on memetracker and dianping.
As shown in
Fig.~\ref{fig:exp_real_data}, without considering any
mark information, PP-poisson and PP-hawkes are unable to handle the temporal
dynamics well on both Memetracker and Dianping.
MPP can discriminate mark-specific time-cost, leading to
better performance than PPs. In memetracker dataset, although RMTPP
has better performance than PP, it does not overbeat
MSPP-poisson and MSPP-hawkes. In dianping dataset, RMTPP(c) and RMTPP(exp)
achieve better performance than MSPP-hawkes when tolerance $\theta\leq 65$
hours, and also achieve better performance than MPP-poisson when tolerance
$\theta\leq 15$ hours.
It is seen that RNN-TD(c) and RNN-TD(exp) achieve the best performance than all
the baselines in the most cases on two datasets. The improvements achieved by
RNN-TD indicate that our proposed method can well model marked temporal
dynamics by learning mark-specific intensity functions,
while RMTPP share the same intensity function for all the marks.

\begin{table}[t!]
\caption{Case study on event prediction}
\label{tb:cs_event_pred}
\centering
\subtable[one specific event sequence prediction on memetraker]{
\label{tb:cs_event_pred_meme}
\resizebox{0.9\textwidth}{!}{
\begin{tabular}{ccrrr}
\hline
\multicolumn{2}{c}{$i$-th event: mark,time(mins)}  & 1th & 2nd & 3rd \\\hline
\multirow{3}{*}{RMTPP} & c\#1 & Europe debt, 22.32 & Europe debt, 12.29 & Europe
debt, 60.31 \\
& c\#2 & LinkedIn IPO, 22.32 & Dominique Strauss, 12.29 & Amy Winehouse, 60.31  \\
& c\#3 & Amy Winehouse, 22.32 & LinkedIn IPO, 12.29 & Dominique Strauss, 60.31  \\
\multirow{3}{*}{RNN-TD} & c\#1 & Europe debt, 1.07 & Dominique Strauss, 1.34 &
Dominique Strauss, 3.63
\\
& c\#2 & Dominique Strauss, 0.45 & Europe debt, 1.29 & Europe debt, 3.12  \\
& c\#3 & LinkedIn IPO, 0.44 & LinkedIn IPO, 0.56 & attack,
2.93
\\
\hline
Ground Truth & & Dominique Strauss, 6.37 & attack, 83.78 & attack, 18.18	 
\\\hline
\end{tabular}
}
}
\qquad
\subtable[one specific event sequence prediction on dianping]{
\label{tb:cs_event_pred_dianping}
\resizebox{0.9\textwidth}{!}{
\begin{tabular}{ccrrr}
\hline
\multicolumn{2}{c}{$i$-th event: mark,time(days)} & 1th & 2nd & 3rd \\\hline
\multirow{3}{*}{RMTPP} & c\#1 & bibimbap, 2.34 & bibimbap, 2.90 & Sichuan
cuisine, 3.02
\\
& c\#2 & tea restaurnt, 2.34 & cookies, 2.90 & cookies, 3.02  \\
& c\#3 & Yunnan cuisine, 2.34 & Sushi, 2.90 & tea restaurnt, 3.02  \\
\multirow{3}{*}{RNN-TD} & c\#1 & bibimbap, 2.93 & barbecue, 0.65 &
barbecue, 0.96
\\
& c\#2 & Yunnan cuisine, 0.88 & bibimbap, 0.85 & Sichuan cuisine, 0.81 \\
& c\#3 & bread, 0.92 & Vietnamese cuisine, 0.51 & bread,
0.48
\\
\hline
Ground Truth & & barbecue,0.14 & Sichuan cuisine,1.03 & barbecue,1.06  \\\hline
\end{tabular}
}
}
\end{table}

\subsection{Case Study on Event Prediction}


To explore the capability of event prediction of RNN-TD, we randomly choose one
specific event sequence from memetracker and dianping respectively, and
estimate the next events in the sequence. 
In RNN-TD, we select top 3 events in descending order of event likelihood as
candidates of next event, called c\#1, c\#2 and c\#3. 
In RMTPP,
we choose the most probable mark and expectation time independently and combine
them as the candidates of next event. 
Table~\ref{tb:cs_event_pred} lists the performance of RMTPP and 
RNN-TD.
We can see that the predicted marks on RNN-TD are more accurate and relevant
to ground truth than compared methods on both cases.
Then, we categorize most relevant marks by
empirical knowledge to evaluate the estimated time on mark-specific methods when
marks are mismatched in all 3 candidates. For example, we consider bibimbap and 
barbecue belong to same regional cuisine, and Dominique Strauss is related
to Europe debt. 
In this way, the average error of time prediction
to ground truth for RNN-TD is 34.55 minutes, and the average error is up to
43.19 minutes for RMTPP in the case of Memetrack. In the case of Dianping,
the average error of time prediction
to ground truth for RNN-TD is 1.13 days, and the average error is nearly doubled
to 2.04 days for RMTPP. Indeed, RNN-TD can provide more options according to
possible event predictions which has more general applications, e.g., 
recommendation systems.

%% file: 5-conclusion.tex
\section{Conclusions}
\label{sec:conclusions}

In this paper, we proposed a general model for marked temporal dynamics
modeling. Based on RNN framework, the representation of hidden layer in RNN-TD
learns the history embedding through a deep structure. The generation of marks
and times is dependent on history embedding so that we can avoid strong prior
knowledge on dependency of history. We observe that the generation processes of
next event are significant different with respect to marks. To capture
the dependence between marks and times, we unfolded
the joint probability of mark and time and parameterized the mark
transition probability and mark-specific conditional intensity function based
on history embedding. We evaluate the effectiveness of our proposed model on two
real-world datasets from memetracker and dianping. Experimental results
demonstrate that our model consistently outperforms existing methods at mark
prediction and time prediction tasks. Moreover, we conduct case study on event
prediction demonstrating that our proposed model is well applicable in marked
temporal dynamics modeling.

%% file: pakdd.bbl
\begin{thebibliography}{10}
\providecommand{\url}[1]{\texttt{#1}}
\providecommand{\urlprefix}{URL }

\bibitem{bao2015modeling}
Bao, P., Shen, H.W., Jin, X., Cheng, X.Q.: Modeling and predicting popularity
  dynamics of microblogs using self-excited hawkes processes. In: Proceedings
  of the 24th International Conference on World Wide Web. pp. 9--10. ACM (2015)

\bibitem{barzel2015constructing}
Barzel, B., Liu, Y.Y., Barab{\'a}si, A.L.: Constructing minimal models for
  complex system dynamics. Nature communications  6 (2015)

\bibitem{cao2010detecting}
Cao, L., Ou, Y., Yu, P.S., Wei, G.: Detecting abnormal coupled sequences and
  sequence changes in group-based manipulative trading behaviors. In:
  Proceedings of the 16th ACM SIGKDD international conference on Knowledge
  discovery and data mining. pp. 85--94 (2010)

\bibitem{chauvin1995backpropagation}
Chauvin, Y., Rumelhart, D.E.: Backpropagation: theory, architectures, and
  applications. Psychology Press (1995)

\bibitem{CranePNAS08}
Crane, R., Sornette, D.: Robust dynamic classes revealed by measuring the
  response function of a social system. Proceedings of the National Academy of
  Sciences  105(41),  15649--15653 (2008)

\bibitem{DuKDD2016}
Du, N., Dai, H., Trivedi, R., Upadhyay, U., Gomez-Rodriguez, M., Song, L.:
  Recurrent marked temporal point processes: Embedding event history to vector.
  In: Proceedings of the 22nd ACM SIGKDD International Conference on Knowledge
  Discovery and Data Mining. pp. 1555--1564. KDD '16, New York, NY, USA (2016)

\bibitem{engle1998autoregressive}
Engle, R.F., Russell, J.R.: Autoregressive conditional duration: a new model
  for irregularly spaced transaction data. Econometrica pp. 1127--1162 (1998)

\bibitem{gao2015modeling}
Gao, S., Ma, J., Chen, Z.: Modeling and predicting retweeting dynamics on
  microblogging platforms. In: Proceedings of the 8th ACM International
  Conference on Web Search and Data Mining. pp. 107--116 (2015)

\bibitem{gomez2013modeling}
Gomez-Rodriguez, M., Leskovec, J., Sch{\"o}lkopf, B.: Modeling information
  propagation with survival theory. In: Proceedings of the 30th International
  Conference on Machine Learning. pp. 666--674 (2013)

\bibitem{gunawardana2016universal}
Gunawardana, A., Meek, C.: Universal models of multivariate temporal point
  processes. In: Proceedings of the 19th International Conference on Artificial
  Intelligence and Statistics. pp. 556--563 (2016)

\bibitem{henaff2016orthogonal}
Henaff, M., Szlam, A., LeCun, Y.: Orthogonal rnns and long-memory tasks. arXiv
  preprint arXiv:1602.06662  (2016)

\bibitem{isaacson1976markov}
Isaacson, D.L., Madsen, R.W.: Markov chains, theory and applications, vol.~4.
  Wiley New York (1976)

\bibitem{kingma2015method}
Kingma, D.P., Adam, J.B.: Adam: A method for stochastic optimization. In:
  International Conference on Learning Representation (2015)

\bibitem{LeskovecKDD2009}
Leskovec, J., Backstrom, L., Kleinberg, J.: Meme-tracking and the dynamics of
  the news cycle. In: Proceedings of the 15th ACM SIGKDD International
  Conference on Knowledge Discovery and Data Mining. pp. 497--506. KDD '09, New
  York, NY, USA (2009)

\bibitem{liuAAAI2016}
Liu, Q., Wu, S., Wang, L., Tan, T.: Predicting the next location: A recurrent
  model with spatial and temporal contexts  (2016)

\bibitem{vaz2013self}
Vaz~de Melo, P.O.S., Faloutsos, C., Assun{\c{c}}{\~a}o, R., Loureiro, A.: The
  self-feeding process: a unifying model for communication dynamics in the web.
  In: Proceedings of the 22nd international conference on World Wide Web. pp.
  1319--1330 (2013)

\bibitem{mikolov2013efficient}
Mikolov, T., Chen, K., Corrado, G., Dean, J.: Efficient estimation of word
  representations in vector space. arXiv preprint arXiv:1301.3781  (2013)

\bibitem{MikolovNIPS2013}
Mikolov, T., Sutskever, I., Chen, K., Corrado, G.S., Dean, J.: Distributed
  representations of words and phrases and their compositionality. In: Advances
  in Neural Information Processing Systems 26. pp. 3111--3119 (2013)

\bibitem{perozzi2014deepwalk}
Perozzi, B., Al-Rfou, R., Skiena, S.: Deepwalk: Online learning of social
  representations. In: Proceedings of the 20th ACM SIGKDD international
  conference on Knowledge discovery and data mining. pp. 701--710 (2014)

\bibitem{pinto2013using}
Pinto, H., Almeida, J.M., Gon{\c{c}}alves, M.A.: Using early view patterns to
  predict the popularity of youtube videos. In: Proceedings of the 6th ACM
  international conference on Web search and data mining. pp. 365--374 (2013)

\bibitem{prechelt1998automatic}
Prechelt, L.: Automatic early stopping using cross validation: quantifying the
  criteria. Neural Networks  11(4),  761--767 (1998)

\bibitem{shen2014modeling}
Shen, H.W., Wang, D., Song, C., Barab{\'a}si, A.L.: Modeling and predicting
  popularity dynamics via reinforced poisson processes. arXiv preprint
  arXiv:1401.0778  (2014)

\bibitem{Szabo2010Commun}
Szabo, G., Huberman, B.A.: Predicting the popularity of online content. Commun.
  ACM  53(8),  80--88 (Aug 2010)

\bibitem{tankov2003financial}
Tankov, P.: Financial modelling with jump processes, vol.~2. CRC press (2003)

\bibitem{tibshirani1996regression}
Tibshirani, R.: Regression shrinkage and selection via the lasso. Journal of
  the Royal Statistical Society. Series B (Methodological) pp. 267--288 (1996)

\bibitem{voorhees1999trec}
Voorhees, E.M.: The trec8 question answering track report. In: Text REtrieval
  Conference (1999)

\end{thebibliography}
